\documentclass[conference]{IEEEtran}
\IEEEoverridecommandlockouts
\usepackage{cite}
\usepackage{amsmath,amssymb,amsfonts}
\usepackage{algorithmic}
\usepackage{graphicx}
\usepackage{textcomp}
\usepackage{xcolor}
\usepackage[numbers]{natbib}
\usepackage{url}
\usepackage{hyperref}
\def\BibTeX{{\rm B\kern-.05em{\sc i\kern-.025em b}\kern-.08em
    T\kern-.1667em\lower.7ex\hbox{E}\kern-.125emX}}
\begin{document}

\title{Retinal–Lipidomics Associations as Candidate Biomarkers for Cardiovascular Health\\
%{\footnotesize \textsuperscript{*}Note: Sub-titles are not captured in Xplore and
%should not be used}
%\thanks{Identify applicable funding agency here. If none, delete this.}
}
\author{
    \IEEEauthorblockN{Inamullah\IEEEauthorrefmark{1}, Imran Razzak\IEEEauthorrefmark{2}, Shoaib Jameel\IEEEauthorrefmark{1}}
    \IEEEauthorblockA{\IEEEauthorrefmark{1}Electronics and Computer Science, University of Southampton, Southampton SO18 1PB, United Kingdom}
    \IEEEauthorblockA{\IEEEauthorrefmark{2}Department of Computational Biology, Mohamed bin Zayed University of Artificial Intelligence, Abu Dhabi, UAE}
    \thanks{Corresponding author: Inamullah (email: i1n23@soton.ac.uk)} % Optional line
}
\maketitle
\begin{abstract}
Retinal microvascular imaging is increasingly recognised as a non-invasive method for evaluating systemic vascular and metabolic health. However, the association between lipidomics and retinal vasculature remains inadequate. This study investigates the relationships between serum lipid subclasses, free fatty acids (FA), diacylglycerols (DAG), triacylglycerols (TAG), and cholesteryl esters (CE), and retinal microvascular characteristics in a large population-based cohort. Using Spearman correlation analysis, we examined the interconnection between lipid subclasses and ten retinal microvascular traits, applying the Benjamini-Hochberg false discovery rate (BH-FDR) to adjust for statistical significance. 

Results indicated that FA were linked to retinal vessel twistiness, while CE correlated with the average widths of arteries and veins. Conversely, DAG and TAG showed negative correlations with the width and complexity of arterioles and venules. These findings suggest that retinal vascular architecture reflects distinct circulating lipid profiles, supporting its role as a non-invasive marker of systemic metabolic health. This study is the first to integrate deep-learning (DL)-derived retinal traits with lipidomic subclasses in a healthy cohort, thereby providing insights into microvascular structural changes independent of disease status or treatment effects.
\end{abstract}

\begin{IEEEkeywords}
Retinal Imaging, Lipidomics, Oculomics, Biomarkers,  Cardiovascular Diseases
\end{IEEEkeywords}
\section{Introduction}

Cardiovascular disease (CVD) is the leading cause of morbidity and mortality worldwide, accounting for nearly 18 million deaths annually \cite{b1}. Early identification of individuals at high risk is critical to preventing major adverse cardiovascular events. Traditional risk scores such as the Framingham Risk Score or Atherosclerotic Cardiovascular Disease (ASCVD) calculators incorporate blood pressure, cholesterol levels, diabetes status, and smoking history. However, they often do not detect subclinical microvascular changes that can occur before obvious signs of the disease appear \cite{b2}. 

The retinal vasculature offers a unique opportunity for non-invasive observation of systemic health due to the structural and functional similarities with other vital organs like the heart and brain. As a model for early vascular changes, the retina is particularly relevant in the study of cardiovascular and metabolic diseases. Advances in fundus photography (FP) and computational tools, such as AutoMorph \cite{b3}, enable precise extraction of metrics from retinal images, including vessel calibre (VC), tortuosity, fractal dimension (FD), and bifurcation geometry. These retinal biomarkers are linked with conditions such as hypertension, type 2 diabetes (T-2D), and cognitive decline,  catalysing the emergence of oculomics\cite{b4}. Evidence suggests that alterations in retinal vessels reflect broader vascular dysfunction, providing opportunities for improved risk assessment, early intervention, and personalised disease monitoring\cite{b5}.

The accessibility of retinal imaging and the strong public participation in eye screening programs further support to use as a non-invasive tool for cardiovascular disease assessment \cite{b4}. For example, Arnould et al. \cite{b6} utilised the Singapore I Vessel Assessment software on images obtained from both optical coherence tomography angiography and FP to predict CVD risk in a clinical cohort of 144 patients. Kim et al. \cite{b7} analysed a larger dataset combining retinal images with diagnostic health information (hypertension, diabetes, smoking status), with age prediction emerging as a dominant factor. Similarly, studies by \cite{b8,b9,b10} reported significant associations between fundus-derived features and CVD risk factors. 

The AutoMorph framework addresses the expertise limitation in manually extracting retinal phenotypes by enabling fully automated and standardised quantification of retinal vascular architecture at scale. Recent studies \cite{b11,b12} have advanced the concept of oculomics, highlighting the eye as a non-invasive indicator of systemic pathophysiology due to shared embryological and vascular traits. However, findings rely on image-based predictive models and overlook the biochemical mechanisms of vascular alterations. Notably, recent findings \cite{b2} indicate that retinal vasculometry independently predicts risks associated with stroke, cardiovascular mortality, and myocardial infarction, establishing vascular morphology as a circulatory biomarker independent of traditional clinical assessments.

Parallel to these developments, lipidomics, the comprehensive profiling of lipid species, has significantly advanced our understanding of metabolic dysregulation in disease. Traditional clinical lipid measures such as total cholesterol, HDL-C, and LDL-C offer limited resolution. In contrast, advanced techniques like ultra-high-performance liquid chromatography coupled with high-resolution mass spectrometry (UHPLC-HRMS) enable the quantification of hundreds of lipid species across structural subclasses. Key subclasses include ceramides (Cer) from sphingolipids; DAG and TAG from glycerolipids; and phosphatidylcholines (PC), ether-linked phosphatidylcholines (PCE), phosphatidylethanolamines (PE), and lysophosphatidylcholines (LPC) from the glycerophospholipid class. In addition, CE and FA are implicated in lipid transport and energy metabolism.  Collectively, these lipid species contribute to membrane structure, intracellular signalling, lipid transport, inflammation, and energy homeostasis, all of which are pathways directly implicated in atherosclerosis, vascular dysfunction, and insulin resistance \cite{b13,b14,b15}. 

On the metabolic side, several studies have highlighted the prognostic value of lipidomic biomarkers. The research carried out at \cite{b16,b17,b18} found that lipid subclasses, particularly DAG, TAG, PC, Cer, and CE, were significantly associated with cardiovascular outcomes, including plaque burden and coronary events, though these were typically assessed alongside traditional risk factors like LDL-C without vascular imaging. Laaksonen et al. \cite{b19} further demonstrated that plasma ceramides predicted cardiovascular death independently of cholesterol measures, suggesting a distinct biological axis of vascular risk. 

Despite this, no prior study has integrated lipidomics with deep-learning retinal phenotyping. While some attempts have combined imaging and omics, they remain limited in scope or focused on diseased populations. For example, DL models trained on fundus photographs predict lipid levels or diabetic status  \cite {b20,b21} , but lack mechanistic interpretability and often use mixed clinical datasets. Bridging these two domains, retinal vascular imaging and high-resolution lipid profiling, offers a novel, non-invasive pathway for identifying early signatures of systemic vascular adaptation and metabolic dysfunction.

In this study, we investigate associations between grouped serum lipid subclass averages and quantitative retinal microvascular features in a large-scale, population-based cohort. By focusing on biologically grouped lipid categories (FA, DAG, TAG, CE) and retinal characteristics derived via the AutoMorph pipeline, our objective is (i) to identify statistically significant lipid-retina associations, (ii) to determine which retinal features are most sensitive to lipid variation, (iii) to preserve natural biological variability by matching participants across lipidomic and retinal datasets, thus minimizing the influence of potential confounding factors, and (iv) to assess whether the retinal vasculature reflects alterations in circulating lipid composition. Our findings offer new insights into the role of specific lipid pathways in microvascular architecture and reinforce the value of the eye as a non-invasive indicator of cardiometabolic health.

\begin{figure}[ht]
    \centering
         \includegraphics[width=\linewidth, height = 9.0 cm]{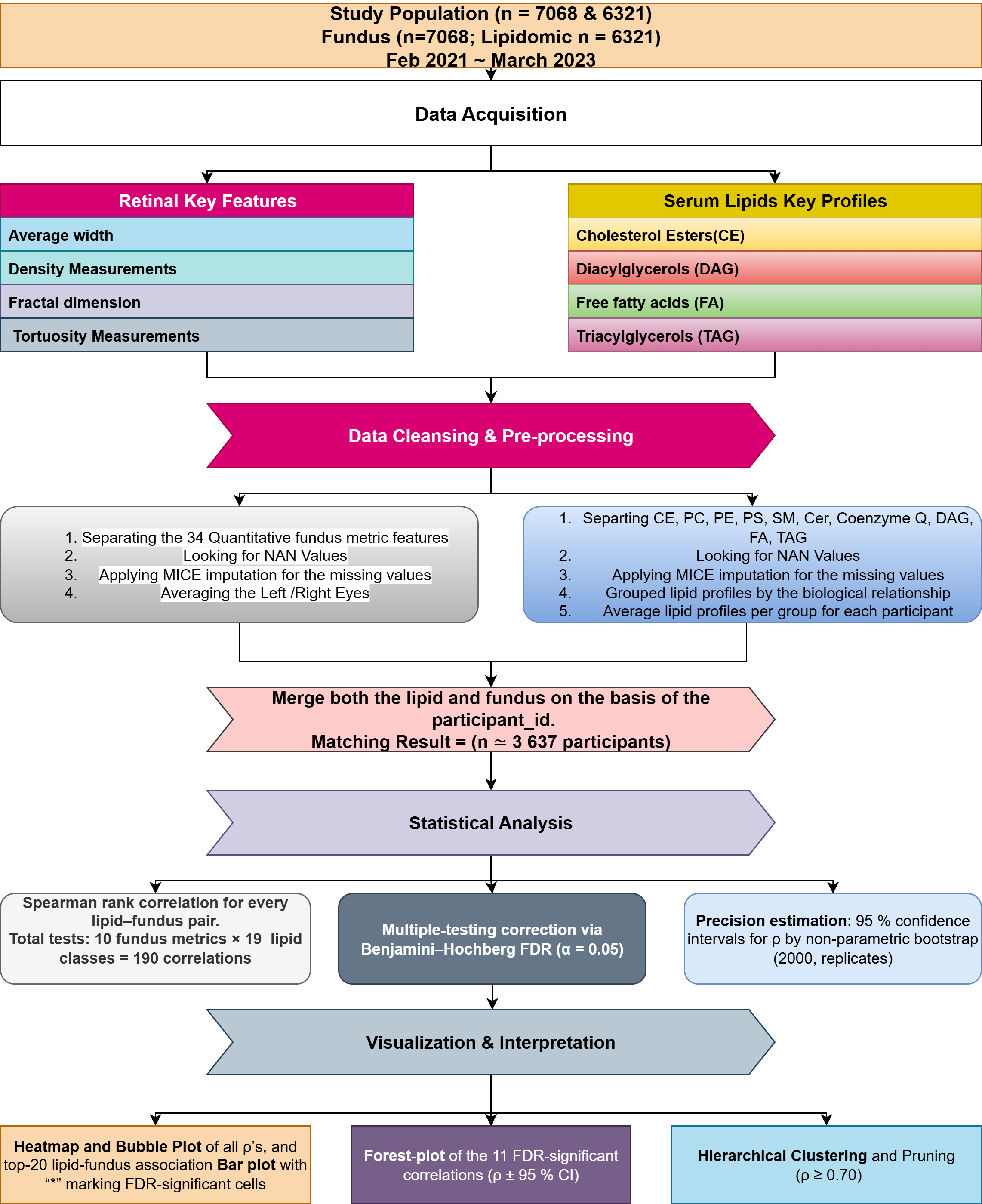}
  \caption{Overview of the study pipeline linking AI-derived retinal vascular traits with serum lipidomics to explore microvascular signatures of cardiometabolic health. This integrative framework enables the identification of non-invasive vascular markers potentially relevant to early cardiovascular risk stratification.}
    \label{fig:1}
\end{figure}
\begin{table}[htbp]
\caption{Baseline Characteristics of the Study Population}
\label{Tab-1}
\centering
\tiny
\setlength{\tabcolsep}{11pt}
\renewcommand{\arraystretch}{0.9}
\begin{tabular}{|l|c|c|c|}
\hline
\textbf{Features} & \textbf{Male} & \textbf{Female} & \textbf{Total} \\
\hline
\textbf{Fundus} & (n= 3443) (48.7\%) & (n= 3625) (51.3\%) & n=7068 \\
\hline
%\multicolumn{4}{|l|}{\textbf{Fundus (n = 7068)}} \\
%\hline
Age (years) & 52.17 ± 7.84 & 53.08 ± 7.87 & 52.64 ± 7.87 \\
Artery Width (px) & 18049.76 ± 1269.98 & 18549.11 ± 1255.81 & 18305.19 ± 1287.08 \\
Artery Density & 0.0391 ± 0.0044 & 0.0390 ± 0.0047 & 0.0391 ± 0.0045 \\
Artery Tortuosity & 0.6929 ± 0.0326 & 0.7010 ± 0.0337 & 0.6971 ± 0.0334 \\
Vein Width (px) & 19198.07 ± 1231.63 & 19619.07 ± 1251.84 & 19413.99 ± 1259.65 \\
Vein Density & 0.0500 ± 0.0042 & 0.0498 ± 0.0044 & 0.0499 ± 0.0043 \\
Vein Tortuosity & 0.7038 ± 0.0246 & 0.7080 ± 0.0250 & 0.7060 ± 0.0249 \\
\hline
\textbf{Lipidomics} & (n= 2944) (46.6\%) & (n= 3377) (53.4\%)  & n=6321 \\
\hline
%\multicolumn{4}{|l|}{\textbf{Lipidomics (n = 6321)}} \\
%\hline
Age (years) & 51.26 ± 8.11 & 52.19 ± 8.05 & 51.76 ± 8.09 \\
CE(22:6) & 0.8892 ± 0.3524 & 1.0128 ± 0.3314 & 0.9556 ± 0.3468 \\
Cer(d18:0/16:0) & 0.3015 ± 0.2206 & 0.2706 ± 0.2132 & 0.2842 ± 0.2169 \\
Cer(d18:1/24:0) & 1.0004 ± 0.1305 & 1.0030 ± 0.1217 & 1.0017 ± 0.1254 \\
CoQ10 & 1.0053 ± 0.2026 & 0.9692 ± 0.1991 & 0.9859 ± 0.2015 \\
DAG(32:1) & 0.0477 ± 0.0965 & 0.0290 ± 0.0653 & 0.0374 ± 0.0813 \\
DAG(36:2) & 0.2468 ± 0.1303 & 0.2050 ± 0.1046 & 0.2243 ± 0.1190 \\
FA(22:0) & 0.1900 ± 0.0707 & 0.1743 ± 0.0583 & 0.1816 ± 0.0648 \\
GlcCer(d18:1/24:1) & 0.3589 ± 0.0926 & 0.3663 ± 0.0968 & 0.3629 ± 0.0950 \\
LPC(18:2) & 0.8892 ± 0.2970 & 0.8614 ± 0.2937 & 0.8742 ± 0.2955 \\
PC(32:1) & 1.0533 ± 0.5342 & 1.1672 ± 0.5303 & 1.1143 ± 0.5351 \\
PCE(34:0) & 0.3588 ± 0.0648 & 0.3912 ± 0.0692 & 0.3761 ± 0.0691 \\
PE(36:3) & 0.4067 ± 0.2063 & 0.4283 ± 0.2085 & 0.4183 ± 0.2077 \\
PI(38:5) & 0.8432 ± 0.2075 & 0.8816 ± 0.2198 & 0.8639 ± 0.2151 \\
PS(38:3) & 0.7524 ± 0.1503 & 0.8043 ± 0.1590 & 0.7802 ± 0.1572 \\
TAG(47:1) & 0.2791 ± 0.2182 & 0.2670 ± 0.2057 & 0.2726 ± 0.2116 \\
TAG(58:10) & 1.1926 ± 0.5455 & 1.1307 ± 0.5354 & 1.1593 ± 0.5409 \\
TAG(62:14) & 0.5258 ± 0.5546 & 0.5565 ± 0.5926 & 0.5423 ± 0.5755 \\
\hline
\end{tabular}

\vspace{1mm}
\begin{minipage}{\columnwidth}
\footnotesize
\textbf{Note:} Values represent mean ± standard deviation for individual baseline fundus and lipidomic features in the study cohort. Fundus traits are measured in pixels (e.g., vessel width) or unitless indices (e.g., vessel density, tortuosity). Lipid features are expressed as log$_{10}$-transformed relative intensities obtained via UHPLC-ESI-HRMS \cite{b22, b23}. These values reflect original cohort distributions prior to multivariate imputation, subclass aggregation, or feature pruning. Further preprocessing details are provided in the Methods section.
\end{minipage}
\end{table}

\section{Methodology}
This study employs a structured analytical pipeline to assess associations between AI-derived retinal vascular features and circulating lipidomic profiles in a healthy cohort. The workflow (Fig.~\ref{fig:1}) encompasses data acquisition, preprocessing, participant matching, dataset fusion, statistical modelling, and interpretation.

From a longitudinal cohort study and a biobank of over 13,000 participants available at \href{https://knowledgebase.pheno.ai/participant_journey.html}{Pheno.AI Participant Journey}, two subcohorts were identified: 7,068 with high-quality fundus imaging and 6,321 with serum lipidomics. After matching and data curation, 3,637 individuals with complete datasets were retained for analysis.Retinal vascular phenotypes were extracted using an automated DL pipeline, while lipidomic profiles were acquired using UHPLC-ESI-HRMS. This framework focuses on linking detailed lipid subclasses to quantitative retinal traits, moving beyond traditional cardiovascular markers. The analysis captures molecular–phenotypic associations suggestive of subclinical vascular changes that may precede disease onset. Statistical assumptions were assessed, including normality and variance checks, followed by FDR correction and bootstrap-based non-parametric correlation estimation. This analytical framework may contribute to the advancement of oculomics by linking molecular lipid signatures with microvascular phenotypes, supporting early detection of vascular ageing and non-invasive biomarker discovery in cardiometabolic research.

\subsection{Study Population and Data Sources}

This cross-sectional analysis utilised data from the Human Phenotype Project (HPP) 10K cohort \cite {b24} , involving 7,068 and 6,321 adult participants from retinal microvasculature and serum lipidomics datasets, respectively \cite {b25,b26}, acquired through the Trusted Research Environment (TRE) \href{https://knowledgebase.pheno.ai/platform_tutorial.html}{Sourced}. Bilateral fundus images were captured via the Icare DRSplus confocal imaging system, providing a 45° macula-centred view without pupil dilation. Fasting venous blood samples underwent untargeted lipidomic profiling through UHPLC-ESI-HRMS. Lipid identification and quantification were conducted using the \href{https://knowledgebase.pheno.ai}{Pheno.AI} lipidomics pipeline (008-serum\_lipidomics), which standardises lipid intensity measures for statistical analysis \cite{b23,b26}. 

Demographic and clinical data, including age, sex, and identifiers, were collected, ensuring participant confidentiality in accordance with the Declaration of Helsinki. The research protocol received Institutional Review Board approval from the Weizmann Institute of Science. The 10K cohort aimed to explore variations in disease susceptibility, clinical phenotypes, and therapeutic responses, contributing to the development of risk prediction models for chronic conditions such as T-2D and CVD \cite{b24}. Given the rising incidence of cardiovascular disease, this work seeks to identify noninvasive surrogate markers for early detection of related conditions.

\subsection{Retinal Image Processing and Feature Extraction}

Retinal fundus images were processed using the open-source AutoMorph framework, developed by Zhou et al.~\cite{b3}. This pipeline, based on DL, is designed for large-scale retinal image analysis and includes modules for image preprocessing, quality grading, anatomical segmentation, and quantitative measurement of vascular morphology. In this study,36 microvascular characteristics were extracted and categorised into the four primary categories for both arteries and veins: Fractal dimension, vessel density, average width, and tortuosity.

FD quantifies the geometric complexity of retinal vasculature, computed using the Minkowski–Bouligand box-counting method~\cite{b27}. Vessel density refers to the proportion of the image area occupied by segmented arteries or veins. Average width is calculated as the total vessel area divided by vessel length and expressed in pixels, where 1 pixel $\approx$ 4.3~\textmu m. Tortuosity metrics capture the curvature and angular deviation of vessel segments, assessed separately for arteries and veins~\cite{b28,b29}. Tortuosity and vessel density are expressed in inverse pixels and as unitless ratios, and FD is unitless by definition. All measurements align with standards used in large-scale retinal imaging studies of the 10K project~\cite{b22}. 

Participants in this study were not diagnosed with any ocular pathology; the focus was solely on quantifying vascular features such as segmental veins and arteries, without considering clinical interpretations of optometric abnormalities such as age-related macular degeneration, microaneurysms, and glaucoma. 

\begin{figure}[ht]
    \centering
         \includegraphics[width=\linewidth]{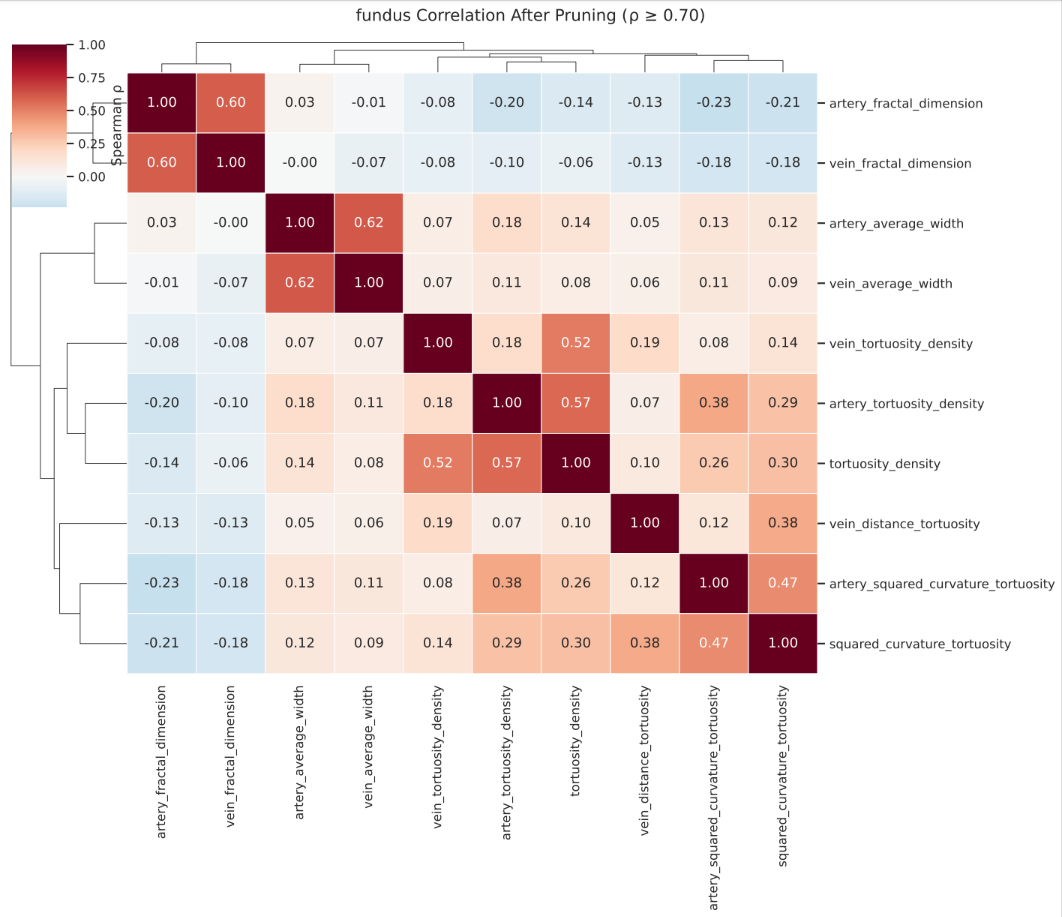}
  \caption{Correlation matrix of fundus microvascular features after collinearity pruning ($\rho$ $\geq$ 0.70).This heatmap shows Spearman correlations between fundus features after removing highly collinear variables. The goal is to retain a set of relatively independent features for downstream analysis. Colour intensity indicates the strength and direction of the correlation.}
    \label{fig:2}
\end{figure}

The bilateral eye fundus images were averaged to produce various main objective phenotypic features, resulting in 18 retinal characteristics per participant. These 18 features underwent hierarchical clustering as illustrated in the Fig \ref{fig:2}, which was reduced to a threshold of ($\rho$ $\geq$ 0.70). Collinear traits were penalised, resulting in a dataset of 7068 participants with 10 features each, and were subsequently used in statistical analyses.

\subsection{Lipidomics Data and Feature Grouping}

Serum lipidomics profiling was conducted using UHPLC-ESI-HRMS. As a subfield of metabolomics, lipidomics focuses on the comprehensive analysis of lipids involved in essential physiological processes such as hormonal regulation, energy storage, membrane integrity, and cellular signalling. These roles make circulating lipids critical biomarkers for systemic and age-related diseases, particularly cardiovascular disease and stroke \cite{b16,b17}. The lipid species were quantified and annotated by the HPP team using established reference libraries and standardised quality control protocols. Comprehensive details on sample preparation, measurement procedures, and data processing workflows are available through the referenced resources \cite{b23,b26}. 

The data, accessed through the TRE, were organised into a structured reference data dictionary encompassing several distinct lipid subclasses. These included: 22:6\_Cholesteryl\_ester, acca\_, Cer\_, coenzyme\_q10, dag\_, fa\_, gsl\_, glccer\_, laccer\_, lysopc\_, pc\_, pce\_, pe\_, pg\_, pi\_, ps\_, sm\_, tag\_, and lysope\_. Each subclass contained multiple molecular species, differing in fatty acyl chain length and degree of unsaturation. For example, the Cer subclass comprised species such as Cer(d18:1/16:0), Cer(d18:1/18:0), and Cer(d18:0/16:0), each defined by specific combinations of sphingoid bases and fatty acid chains.

We selected biologically relevant lipid groups, TAG, DAG, CE, FA, PC, SM, and Cer—totalling 187 species for correlation analysis with retinal microvascular traits, based on prior evidence linking specific lipid subclasses to CV risk \cite{b13,b17,b18}. To enhance interpretability, we computed mean concentrations within each subclass. 

Recognising the potential bias from missing values $\approx$ 20\%, due to lipidomics profiling, we employed Multivariate Imputation by Chained Equations (MICE) to accurately impute these values while preserving the dataset's multivariate structure. Post-imputation, lipid species were aggregated and filtered using a Spearman correlation threshold of ($\rho$ $\geq$ 0.70) to reduce redundancy and collinearity, thereby boosting variability and statistical power in the subsequent analyses.
%Based on prior evidence linking specific lipid subclasses to cardiovascular risk \cite{b18,b16,b19}, we selected biologically relevant lipid groups for downstream analysis. These included TAG, DAG, CE, FA, PC, SM, and Cer, yielding a total of 187 lipid species across these groups. To reduce dimensionality and improve interpretability, we computed the mean concentration of molecular species within each subclass, generating composite features for use in correlation analysis with retinal microvascular traits. However, due to the inherent sparsity and quality-filtering steps involved in lipidomics profiling, some species contained missing values (NaNs), which can introduce bias and instability in statistical modelling.

%To address this, we applied Multivariate Imputation by Chained Equations (MICE), a robust technique that preserves the underlying distribution and captures complex interdependencies among variables. This method ensured accurate imputation of missing values while maintaining the multivariate structure of the dataset. Following imputation, lipid species were aggregated into their respective subclasses, and further pruned by applying a Spearman correlation threshold of ($\rho$ $\geq$ 0.70) (as illustrated in Fig. \ref{fig:3}). This pruning step helped minimise redundancy and collinearity within each lipid class, thereby enhancing the variability and statistical power of subsequent analyses.

\begin{figure}[ht]
    \centering
         \includegraphics[width=\linewidth]{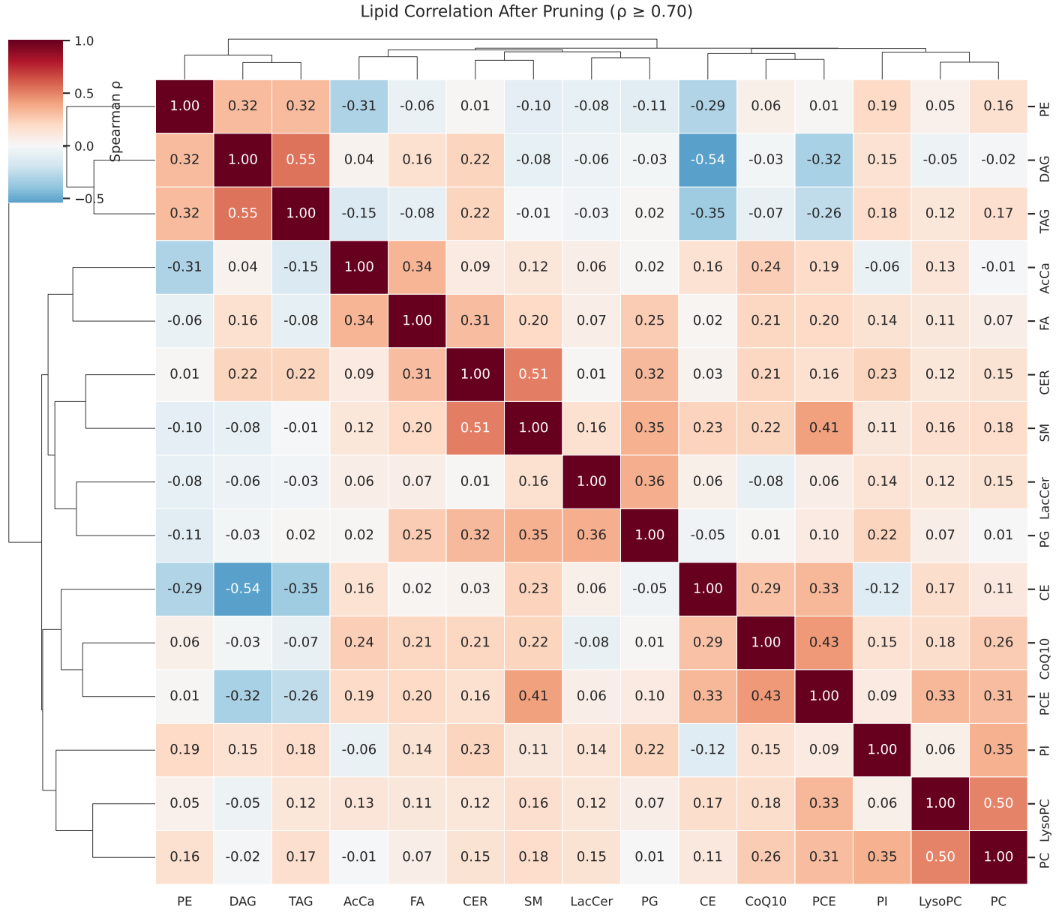}
  \caption{Correlation matrix of lipid subclass features after collinearity pruning ($\rho$ $\geq$ 0.70). Spearman correlation heatmap of averaged lipid subclasses after filtering out highly correlated features. This step reduces multicollinearity and ensures that selected lipid groups represent distinct biological signals. Clustering highlights relationships among remaining lipid classes.}
    \label{fig:3}
\end{figure}

\subsection{Statistical Analysis}

To explore the potential links between circulating lipid profiles and retinal microvascular traits, we conducted a structured statistical analysis designed to handle the complexity and biological variability of both datasets. Before diving into the correlation analysis, we first examined the distribution of the variables. Using tools such as the Shapiro–Wilk test and Q–Q plots, we found that most lipid and fundus features showed clear departures from normality. This included skewed distributions and heavy tails, which are common in biological datasets. Moreover, visual inspection of scatter plots and results from Levene's test revealed that many feature pairs lacked linear relationships and showed unequal variance across the range of values. Given these violations of key assumptions required for traditional Pearson correlation, including normality, linearity, and homoscedasticity, we opted to use Spearman's rank correlation coefficient instead. 

Spearman correlation is a robust, non-parametric alternative that doesn’t assume a normal distribution and is better suited to capturing monotonic patterns, even when relationships are non-linear or influenced by outliers. To ensure the integrity of the analysis, we first addressed the presence of missing values. For lipidomics data, which is known to contain gaps due to instrument detection limits or stringent quality control filters, we applied MICE. This method allowed us to estimate missing values in a way that maintains the overall structure and variability of the data. For the fundus dataset, we averaged left and right eye measurements and reduced further through hierarchical clustering to reduce redundancy and noise, resulting in a consistent set of 10 vessel traits per participant. 

Following these steps, we merged the two datasets using an inner join on participant\_id, keeping only those participants with complete information in both domains. This resulted in a final analysis cohort of 3,637 healthy individuals, offering a clean and consistent foundation for downstream analysis. We then calculated pairwise Spearman correlations between the 19 aggregated lipid subclass features and the 10 fundus traits, resulting in 190 correlation pairs. For each pair, we computed two-sided p-values and applied BH-FDR correction to account for multiple testing, using an adjusted significance threshold of FDR ($\rho$) $<$ 0.05. 

To evaluate the stability and confidence of each correlation estimate, we also implemented bootstrap resampling (2,000 iterations) to compute 95\% confidence intervals (CI). These steps ensured that our findings were not only statistically valid but also reproducible and interpretable. The key associations are presented in both tabular and visual formats, including heatmaps and forest plots, to highlight patterns and biological relevance across the lipid–retina landscape.
\begin{table}[htbp]
\caption{Significant Fundus--Lipid Associations ($FDR < 0.05$)}
\label{Tab-2}
\centering
\tiny
\setlength{\tabcolsep}{11pt}
\renewcommand{\arraystretch}{0.9}
\begin{tabular}{|l|l|c|c|c|}
\hline
\textbf{Fundus Feature} & \textbf{Lipid Feature} & \boldmath{$\rho$} & \textbf{95\% CI} & \textbf{FDR (q)} \\
\hline
artery\_average\_width     & DAG   & -0.14  & [-0.17,\ -0.11] & $4.83\times10^{-15}$ \\
artery\_average\_width     & TAG   & -0.13  & [-0.16,\ -0.10] & $2.62\times10^{-13}$ \\
artery\_average\_width     & CE    &  0.09  & [ 0.06,\  0.12] & $6.90\times10^{-6}$  \\
artery\_average\_width     & PCE   &  0.09  & [ 0.06,\  0.12] & $1.70\times10^{-6}$  \\
vein\_fractal\_dimension   & FA    &  0.08  & [ 0.05,\  0.11] & $7.14\times10^{-5}$  \\
vein\_average\_width       & DAG   & -0.06  & [-0.09,\ -0.03] & $7.00\times10^{-4}$  \\
artery\_fractal\_dimension & TAG   & -0.06  & [-0.09,\ -0.03] & $1.18\times10^{-4}$  \\
vein\_average\_width       & CE    &  0.06  & [ 0.02,\  0.09] & $2.00\times10^{-3}$  \\
squared\_curvature\_tortuosity & FA & -0.05  & [-0.08,\ -0.02] & $3.60\times10^{-2}$  \\
tortuosity\_density        & FA    & -0.05  & [-0.08,\ -0.02] & $3.60\times10^{-2}$  \\
artery\_tortuosity\_density& FA    &  -0.05  & [-0.08,\ -0.02] & $3.60\times10^{-2}$  \\
\hline
\end{tabular}
\end{table}
\section{Results}

This section presents the integrated analysis of AI-derived retinal vascular traits and serum lipidomic profiles within a healthy cohort. A total of 7,068 bilateral fundus images were processed using the AutoMorph pipeline, generating 36 vascular features per participant. After averaging left and right eye values and applying hierarchical clustering to reduce feature redundancy ($\rho \geq 0.70$), 10 representative retinal traits were retained. These were fused with lipidomics data and resulted in 3,637 individuals. 

The results begin with a descriptive summary of the study cohort and dataset structure depicted in the Table.\ref{Tab-1}, and \ref{Tab-2}, followed by correlation analyses between retinal traits and lipid subclasses using FDR-adjusted Spearman analysis. Subsequent findings highlight lipid class–specific associations with microvascular architecture, capturing how variations in circulating lipids align with retinal vascular features.
%This section summarises the analysis of fundus imaging features integrated with serum lipidomics profiles, focusing on four primary objectives. A total of 7,068 bilateral retinal images were processed using the AutoMorph pipeline, extracting 36 vascular features per participant. Left and right eye features were averaged, and hierarchical clustering reduced feature redundancy, resulting in 10 independent retinal traits after removing collinear variables ($\rho$ $\geq$ 0.70). These traits were then matched with lipidomics data from 3,637 healthy individuals, encompassing 187 serum lipid species. The results address: (i) baseline cohort characteristics; (ii) significant lipid–retina correlations identified through Spearman rank analysis with FDR correction; (iii) influential retinal features associated with lipid variation; and (iv) interpretation of these associations in relation to cardiovascular health and circulatory remodelling. Collectively, these findings elucidate the impact of circulating lipids on retinal microvascular architecture, suggesting the eye may serve as an indicator of systemic cardiometabolic health.

\begin{figure}[ht]
    \centering
    \includegraphics[width=\linewidth]{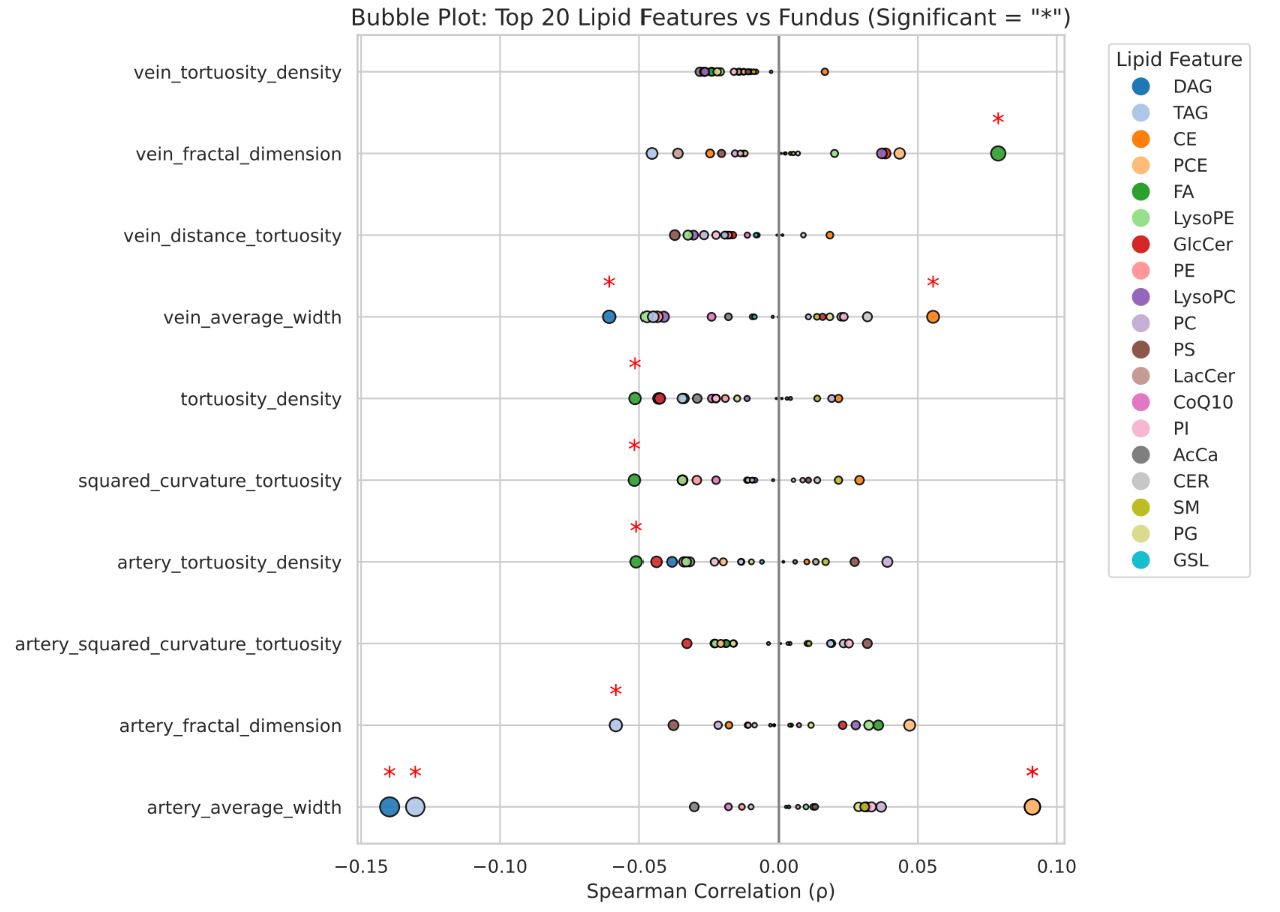}
    \caption{Bubble plot of top 20 lipid features correlated with fundus vascular traits. Each bubble represents a Spearman correlation ($\rho$) between a fundus feature (y-axis) and one of the top 20 lipid subclasses. Bubble size reflects the absolute correlation magnitude, and colour denotes lipid subclass identity. Red asterisks (*) indicate statistically significant correlations after FDR correction.}
    \label{fig:4}
\end{figure}

\subsection{Lipid–Retina Correlations}

The pruned 190 lipid–retina pairs, obtained after FDR-BH multiple correction (q$<$0.05), yielded 11 unique and statistically significant associations. The observed variation in retinal microvascular characteristics across lipid subclasses offers a molecular lens into how circulating lipid profiles relate to objective, quantifiable features of the retina. This cross-modal relationship underpins the potential of oculomics in decoding systemic metabolic health through the eye.

As shown in Fig.~\ref{fig:4}, 19 lipid subclasses exhibited strong correlations with ten fundus features. A vertical black line separates the directions of correlation (positive and negative), while red asterisks indicate statistical significance. Among these, artery average width emerged as the most consistently associated trait, demonstrating robust correlations with DAG, TAG, and PCE, spanning both directions. Vein average width followed closely, showing significant associations particularly with DAG and CE subclasses. Beyond VC, various tortuosity-based measurements, such as squared curvature tortuosity, artery tortuosity density, and tortuosity density, revealed more selective associations, predominantly with FA subclasses. Similarly, the artery and vein fractal dimensions displayed distinct yet parallel relationships with TAG and FA, respectively, reflecting consistent geometric responses to systemic lipid exposure. 

These associations resonate with vascular pathophysiology observed in early stage atherosclerosis and arteriosclerosis, where lipid deposition, oxidative stress, and endothelial dysfunction contribute to narrowing of the lumen, vascular stiffness, and elevated blood pressure \cite{b13,b30,b31,b32}. The patterns uncovered here suggest that retinal imaging biomarkers may offer an early, non-invasive signal of systemic lipid remodelling, an idea we further explore in the following discussion.
\begin{figure}[ht]
    \centering
    \includegraphics[width=\linewidth]{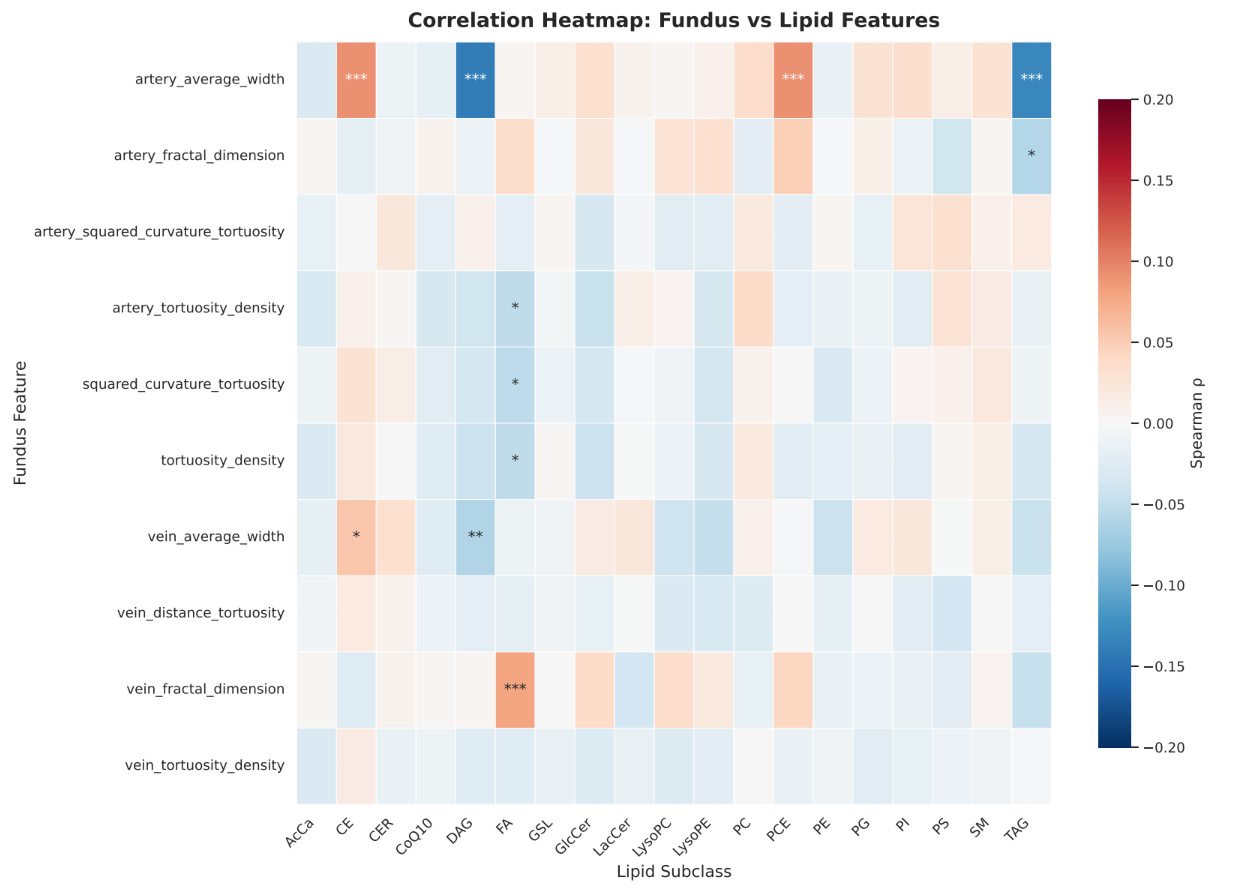}
    \caption{ Heatmap of Spearman correlations between fundus microvascular features and lipid subclasses. Cells are coloured by Spearman correlation coefficient ($\rho$), ranging from $-0.2$ (blue) to $+0.2$ (red). Asterisks indicate statistical significance after FDR correction: $p < 0.05$ (*), $p < 0.01$ (**), and $p < 0.001$ (***). Notable associations include \texttt{artery\_average\_width} with CE, DAG, TAG, and PCE subclasses, and \texttt{vein\_average\_width} with CE and DAG.}
    \label{fig:5}
\end{figure}

\begin{figure}[ht]
    \centering
    \includegraphics[width=\linewidth]{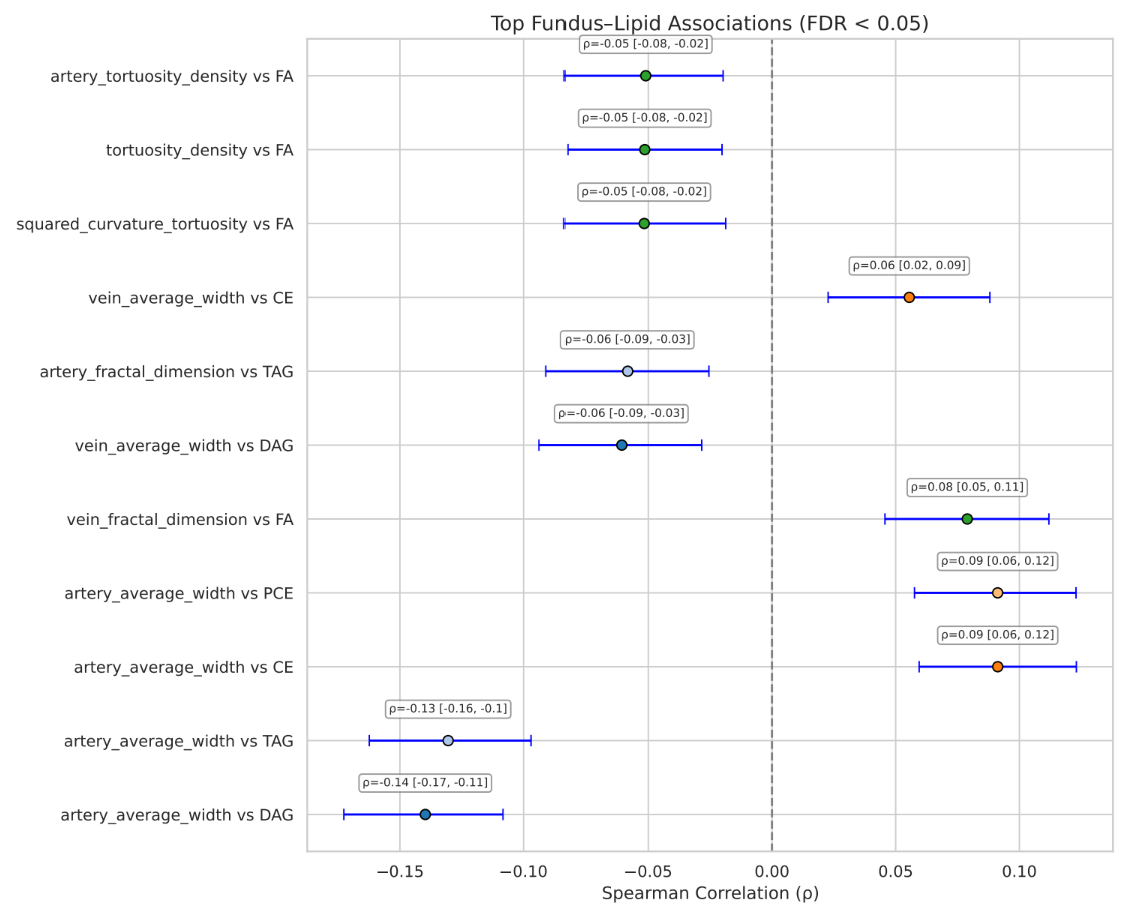}
    \caption{Forest plot of top fundus–lipid associations after FDR correction (q $<$ 0.05). Each point represents a significant Spearman correlation ($\rho$) between a retinal vascular feature and a lipid subclass. Error bars denote 95\% CI based on bootstrapping.}
    \label{fig:6}
\end{figure}

\subsection{Visualization of Associations}

To elucidate and interpret the observed associations, a series of visualisations was employed. The Spearman correlation matrix (Fig.\ref{fig:5}) served as an initial overview, highlighting consistent patterns of lipid–retina associations across vessel types. Statistically significant correlations (q $<$ 0.05) were denoted with asterisks, with prominent clusters involving CE, DAG, and TAG subclasses and arterial width, fractal dimension, and vessel density.

A forest plot (Fig. \ref{fig:6}) detailed the effect sizes and CI for the 11 significant associations, enabling direct comparison of the strength and directionality of individual correlations. Most notably, artery average width appeared as a recurrent fundus trait linked with lipid subclasses implicated in cardiovascular risk, underscoring its potential role as a vascular biomarker. The bubble plot (Fig. \ref{fig:4}) highlighted the top 20 lipid–retina associations, with bubble size indicating correlation magnitude and colour denoting lipid class. Here, FA and CE subclasses were enriched among positively correlated venular traits, while DAG and TAG showed concentrated negative associations with arteriolar features. This visualisation reinforced the vessel-type specificity of lipid impacts and demonstrated potential structural biomarkers detectable through non-invasive imaging.

\section{Discussions}

In this population-based study of 3,637 healthy individuals, we explored how retinal vascular traits reflect systemic lipid variation, offering a non-invasive window into vascular health. Using FDR-adjusted Spearman correlation, eleven significant lipid–retina associations were identified across ten vascular traits. 

A key novel finding was the consistent relationship between lipid subclasses and multiple retinal features, including vascular calibre, tortuosity, and fractal dimension, across both arteries and veins.
Retinal arterial narrowing, indicated by reduced average artery width, was inversely correlated with serum levels of DAG ($\rho = -0.14$, CI: $-0.17$ to $-0.11$) and TAG ($\rho = -0.13$, CI: $-0.16$ to $-0.10$). In contrast, the PCE and CE showed a positive direction ($\rho = +0.09$, CI: $+.06$ to $+0.12$), suggesting that elevated lipid levels may contribute to vascular dilation. 

Beyond calibre, structural geometry also varied with lipid exposure. The geometric complexity of venules, measured by vein fractal dimension, was positively associated with FA ($\rho = +0.08$, CI: $+.05$ to $+0.11$), indicating a potential change in response to lipid-induced stress \cite{b30,b33}.
The most consistent associations were observed between FA ($\rho = -0.05$, CI: $-0.08$ to $-0.02$) and tortuosity-based metrics. The FA levels were linked to reduced tortuosity, measured across artery\_tortuosity\_density, tortuosity\_density, and squared\_curvature\_tortuosity, considering the elevation of dyslipidemia. These findings echo Cheung et al. \cite{b30}, who found that reduced arteriolar tortuosity was independently associated with higher blood pressure and body mass index, hallmarks of cardiovascular risk. From a pathophysiological perspective, vessel straightening may reflect increased stiffness, compromised vascular compliance, or chronic endothelial stress, all possible outcomes of long-term FA elevation \cite{b31, b17}. 

Upon looking at the venous structure, widening is common; however, vasoconstriction is rare. Wong et al \cite{b33} highlight this mechanism that the retinal venular diameter declines with the passage of time, and it is more common in older age than in younger age. In our work, the driving force for this was DAG lipids ($\rho = -0.06$, CI: $-0.09$ to $-0.03$), which may be deposited in the veins and can cause cardiometabolic risk \cite{b13}. The most commonly observed retinal vascular changes include arterial narrowing, along with widening of both arteries and veins. These structural variations may reflect imbalances in vasomotor tone, potentially contributing to vascular dysfunction seen in conditions such as hypertension, thrombus formation, ischemia, and stroke \cite{b32}. 

The glycerolipids (TAG, DAG) are linked to the rarefactions of the arterial tree, while the glycerophospholipids (PCE) and CE are linked to the rupturing and dilations. These three main classes of lipidomics serum profiles cover half of the main features of our study and capture the shared molecular axis that potentially changes the retinal vasculature, and previous work also suggested that these associations are driving factors for mortality and cardiovascular disease outcomes \cite{b13,b16,b34,b31}.
Arterial fractal dimension, an index of vascular branching complexity, was negatively associated with TAG ($\rho = -0.06$, CI: $-0.09$ to $-0.03$), reinforcing evidence that arterial branching becomes simpler and stiffer with age and dyslipidemia \cite{b35,b36}. Likewise, FA’s positive relationship with venular complexity may reflect compensatory dilation or alteration under increased lipid load \cite{b37}. 

To summarise, these microvascular changes may reflect early manifestations of atherogenesis, endothelial dysfunction, or hypertension, conditions that benefit from timely detection and are tightly linked to modifiable risk factors such as blood pressure, obesity, and dyslipidemia \cite{b2,b9}. Collectively, our findings support the hypothesis that the retina serves as a sensitive, non-invasive window into systemic lipid metabolism and vascular health. In particular, many of the retinal features identified, specifically artery width and tortuosity, are not only quantifiable via AI-assisted fundus imaging but also linked with established cardiometabolic parameters. 

These traits warrant further investigation as possible indicators for early risk stratification and monitoring. Furthermore, integrating retinal imaging with lipidomics may enhance the detection of subclinical vascular alterations before the onset of overt disease, supporting scalable screening strategies. This translational relevance underscores the emerging role of retinal biomarkers in precision medicine and supports their potential consideration in future cardiovascular risk prediction frameworks.

\section{Limitations and Future Directions}
This study's findings on lipid-retinal associations are accompanied by several limitations. The cross-sectional design restricts causal inferences regarding serum lipid levels and vascular remodelling, necessitating longitudinal studies to assess whether retinal microvascular alterations precede systemic cardiometabolic events. 

Additionally, while efforts were made to control for confounding factors through participant matching and exclusion of overt disease cases, residual confounders such as diet, genetics, and medication use may still affect lipid and vascular phenotypes. Moreover, the focus on average lipid subclass intensities and aggregate vessel traits might overlook subtle molecular and segmental relationships. Future investigations should consider lipid species-specific effects and regional vessel segmentation via high-resolution imaging. 

Lastly, although AutoMorph improves feature extraction standardisation, external validation across diverse cohorts and imaging systems is crucial for confirming generalizability. Integrating genomic, transcriptomic, and proteomic data with lipidomics and imaging may elucidate deeper biological pathways for precision cardiovascular screening.
\section{Conclusion}
This study is the first to integratively analyse serum lipid subclasses and retinal microvascular morphology in a large, healthy cohort using automated deep learning. We identified significant associations between glycerolipids, phospholipids, cholesterol esters, fatty acids, and changes in retinal vessel narrowing, tortuosity, and dilation. 

Our findings suggest that the retina reflects systemic lipid variation, providing measurable and interpretable subclinical indicators of vascular risk. By matching data via participant\_id and avoiding traditional confounding adjustments, we preserved biological variability and captured a more authentic signal of early vascular changes. This research highlights the potential of integrating retinal imaging with lipidomics as a non-invasive strategy for cardiovascular risk assessment. 
\bibliographystyle{IEEEtran}
\bibliography{Ref}

\end{document}